\documentclass[10pt, a4paper]{article}

\usepackage{lrec2026} % this is the new style
\usepackage{makecell}
\usepackage{multirow}
\usepackage{amsmath,amssymb,mathtools}
\usepackage{dsfont}                
\usepackage[table,xcdraw]{xcolor}
\usepackage{hyperref}
\usepackage{float}
\usepackage{booktabs}

\makeatletter
\def\ps@plain{\ps@empty} % 'plain' 스타일(표지/첫 페이지 등)도 비움
\makeatother
\title{VG-CoT: Towards Trustworthy Visual Reasoning via Grounded Chain-of-Thought}

\name{Byeonggeuk Lim\textsuperscript{1}, Kyeonghyun Kim\textsuperscript{2}, JungMin Yun\textsuperscript{2}  and YoungBin Kim\textsuperscript{1, 2}} 

\address{Graduate School of Advanced Imaging Science, Multimedia \& Film, Chung-Ang University\textsuperscript{1}\\
Department of Artificial Intelligence, Chung-Ang University\textsuperscript{2} \\
         \{banggeuk, khyun8072, cocoro357, ybkim85\}@cau.ac.kr\\}

\abstract{
The advancement of Large Vision-Language Models (LVLMs) requires precise local region-based reasoning that faithfully grounds the model’s logic in actual visual evidence. However, existing datasets face limitations in scalability due to extensive manual annotation and lack of explicit alignment between multi-step reasoning and corresponding image regions, which constrains the evaluation of model trustworthiness. To address these challenges, we propose the Visual Grounding Chain-of-Thought (VG-CoT) dataset, which explicitly links each reasoning step to real visual evidence within the image through a fully automated three-stage pipeline. The pipeline first extracts object- and text-level visual evidence using state-of-the-art detection and OCR models, then generates step-by-step grounded reasoning with GPT-4o, and finally refines the grounding through a rationale-driven open-set detection process. In addition, we introduce a new benchmark that comprehensively evaluates LVLMs reasoning across three complementary dimensions: Rationale Quality, Answer Accuracy, and Reasoning-Answer Alignment. Experiments with representative LVLMs, including LLaVA-1.5 and Qwen2-VL, demonstrate consistent improvements on most evaluation metrics, confirming that VG-CoT effectively enhances trustworthy, evidence-based reasoning while maintaining scalable and cost-efficient dataset construction. The dataset and code will be released publicly upon acceptance to facilitate further research.
 \\ \newline  }

\begin{document}

\maketitleabstract

\section{Introduction}
\label{sec:intro}
The development of Large Language Models (LLMs) has recently led to the rise of Large Vision-Language Models (LVLMs), which simultaneously understand visual and linguistic information \cite{1_zhang2024vision, 2_yin2024survey}. LVLMs have demonstrated outstanding performance in comprehensive, image-level understanding, and the focus of research is gradually expanding toward the ability for precise understanding of local regions within an image \cite{3_NEURIPS2023_9a6a435e, 24_Liu_2024_CVPR, 25_zhuminigpt, 26_wang2024qwen2}. Specifically, the ability to accurately identify specific regions within an image and interpret their spatial and semantic relationships is emerging as a core requirement for performing advanced vision-language tasks such as autonomous driving, robotics, and medical image analysis \cite{4_li-etal-2024-topviewrs, 5_pmlr-v229-zitkovich23a,6_NEURIPS2023_5abcdf8e}.

However, this local region-based reasoning ability is highly dependent on the quality and composition of the training dataset \cite{36_wu2023multimodal,37_chang2024sur}. Currently, most widely used vision-language datasets primarily focus on evaluating comprehensive, image-level understanding, which limits models from learning fine-grained reasoning based on the detailed attributes of individual objects or the relationships between objects \cite{7_chen-etal-2024-measuring, 8_zhou2025perception, 13_Antol_2015_ICCV, 14_Hudson_2019_CVPR}. Although some studies have proposed datasets that include local region information, they still rely on large-scale manual annotation, making it difficult to avoid the constraints of cost and scalability \cite{27_yu2016modeling, 28_Zellers_2019_CVPR, 15_krishna2017visual}. This manual construction method presents a more pronounced limitation, especially in tasks that require complex visual relationships or multi-step reasoning. This suggests the need for a dataset that can support the learning of complex, local region-based reasoning in a scalable and efficient manner.

Existing multimodal Chain-of-Thought (CoT) datasets designed to support multi-step reasoning suffer from a structural limitation: the reasoning process is not explicitly linked to specific visual evidence within the actual image \cite{16_NEURIPS2022_11332b6b, 27_yu2016modeling}. In other words, because the location information (spatial coordinates) of the objects or text within the image that serve as the rationale for the inference is not provided, it is difficult to verify which areas of the image the model's reasoning was based on. This absence of visual evidence leads to a structural limitation in model evaluation regarding whether the model derived the correct answer using the right evidence as its rationale, causing existing benchmarks to only assess the accuracy of the final answer \cite{36_wu2023multimodal, 37_chang2024sur}. Therefore, a dataset where the entire process of reasoning is explicitly aligned with real visual evidence is essential for verifying whether the logic presented by the model is faithful to the actual image evidence \cite{37_chang2024sur, 38_qiu-etal-2024-valor}. This also provides the foundation for a new benchmark capable of systematically measuring not only the accuracy of the final answer but also the accuracy of the rationale and the logical soundness of the reasoning process.

Based on this necessity, this study proposes the Visual Grounding Chain-of-Thought (VG-CoT) dataset. VG-CoT is a dataset constructed by explicitly aligning visual evidence from an image with the reasoning process across various tasks, achieving both efficiency and scalability through an automated three-stage pipeline. (i) First, initial visual evidence is extracted using state-of-the-art object detection and OCR models. (ii) Based on this, step-by-step reasoning is generated using GPT-4o. (iii) Finally, using the generated reasoning process as a clue, visual evidence is more precisely captured through a text-based object detection model. This automated pipeline effectively solves the cost issue associated with manual annotation while contributing to the improvement of the quality of the reasoning process and the reliability of rationale alignment.

Furthermore, this study proposes a new benchmark that goes beyond the existing evaluation methods centered on answer accuracy. This benchmark comprehensively measures Rationale Quality, Answer Accuracy, and Reasoning-Answer Alignment, which represents the relationship between the two metrics. Specifically, it utilizes GPT-4o \cite{11_achiam2023gpt} as an evaluator to precisely assess the reasoning process using detailed metrics: logical coherence, reasoning completeness, and visual evidence utilization. This allows for an in-depth verification of whether the model has reasoned logically based on credible visual evidence, going beyond simply checking if the model provided the correct answer.

The key contributions of this study are summarized as follows:
\begin{enumerate}
    \item We propose the VG-CoT dataset, which explicitly aligns the reasoning process with real visual evidence within the image through an automated three-stage pipeline.
    \item We establish a new benchmark that comprehensively measures Rationale Quality and Reasoning–Answer Alignment, going beyond simple answer accuracy.
    \item We evaluate representative LVLMs using VG-CoT to analyze their capability in utilizing visual evidence, providing insights into the future direction of LVLMs research.
\end{enumerate}

\section{Related Work}
\label{sec:related_work}
\begin{figure*}[t]
    \centering
    \includegraphics[width=1.0\textwidth]{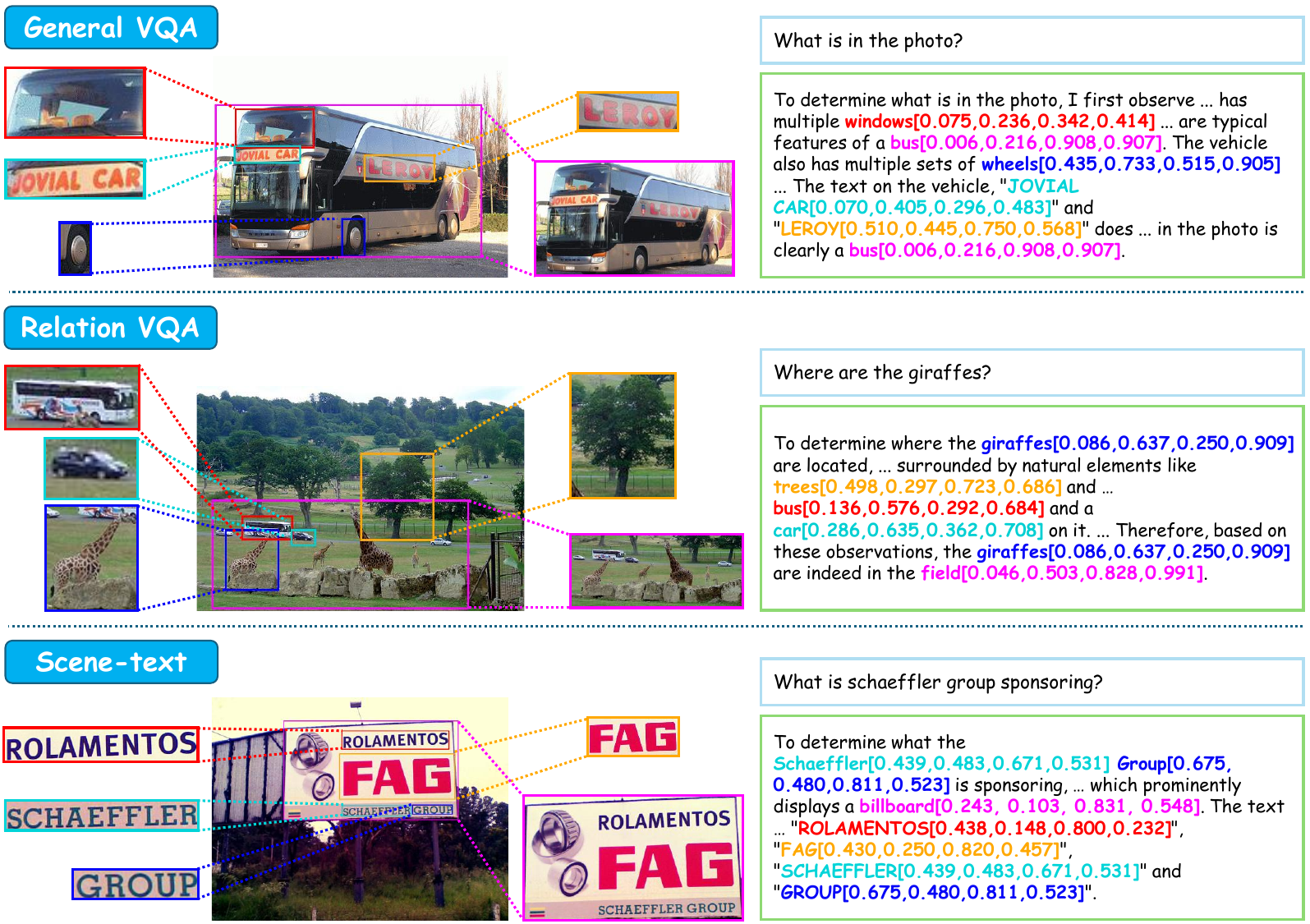}
    \caption{Examples of the Proposed VG-CoT Dataset across Three Task Types.}
\label{fig:figure1}
\end{figure*}

\subsection{Datasets for LVLMs Reasoning}

Early datasets for LVLMs primarily focused on image-wide understanding, which limited their ability to train models for fine-grained, local region-based reasoning \cite{12_lin2014microsoft, 13_Antol_2015_ICCV, 14_Hudson_2019_CVPR}. To overcome this limitation, datasets that explicitly include local region information were proposed, but most relied on manual annotation, a method that requires immense cost and time, thereby restricting data diversity and scalability \cite{15_krishna2017visual, 23_Peng_2024_CVPR, 27_yu2016modeling}. Consequently, there emerged a need for datasets that overcome the limitations of manual annotation and require scalable yet detailed visual understanding.

In line with this necessity, studies have been conducted to encourage models to perform complex reasoning processes \cite{10_NEURIPS2024_0ff38d72, 16_NEURIPS2022_11332b6b, 17_chen-etal-2024-m3cot, 18_Man_2025_CVPR}. ScienceQA \cite{16_NEURIPS2022_11332b6b} aimed to enhance model explainability by providing rationales for questions, and subsequently, M3CoT \cite{17_chen-etal-2024-m3cot} pointed out that ScienceQA was limited to a specific domain and single-step reasoning. M3CoT was thus designed to necessarily require multi-step reasoning across multiple domains, increasing the complexity of inference. These studies marked significant progress in that they explicitly attempted to show the reasoning process through multimodal CoT. However, a limitation shared by both studies is the lack of an explicit link between each reasoning step and specific visual evidence within the image. Grounding the reasoning process in actual visual evidence is essential to verify that the model's logic relies on true image content rather than merely producing plausible text, which fosters trust in the model.

To address this problem of the lack of visual evidence, Visual CoT \cite{10_NEURIPS2024_0ff38d72} attempted to guide the model's focus by adding a single core region bounding box, which is necessary for the question and answer, as CoT information. However, Visual CoT primarily focused on a single region related to the final answer and was biased toward only one type of evidence, either object or text. This still limits its ability to train complex reasoning skills that require the comprehensive referencing of multiple objects and text evidence during a multi-step reasoning process \cite{10_NEURIPS2024_0ff38d72, 16_NEURIPS2022_11332b6b, 17_chen-etal-2024-m3cot}. Consequently, previous studies left limitations including (1) difficulty in achieving scalability due to manual construction, (2) the disconnection between the reasoning process and visual evidence, and (3) insufficient utilization of multiple object and text evidence. To simultaneously solve these issues, this study proposes the VG-CoT dataset, which explicitly aligns the location information of multiple objects and text with each reasoning step through an automated three-stage pipeline, thereby laying the groundwork for complex and trustworthy rationale-based visual reasoning.

\begin{figure*}[t]
    \centering
    \includegraphics[width=1.0\textwidth]{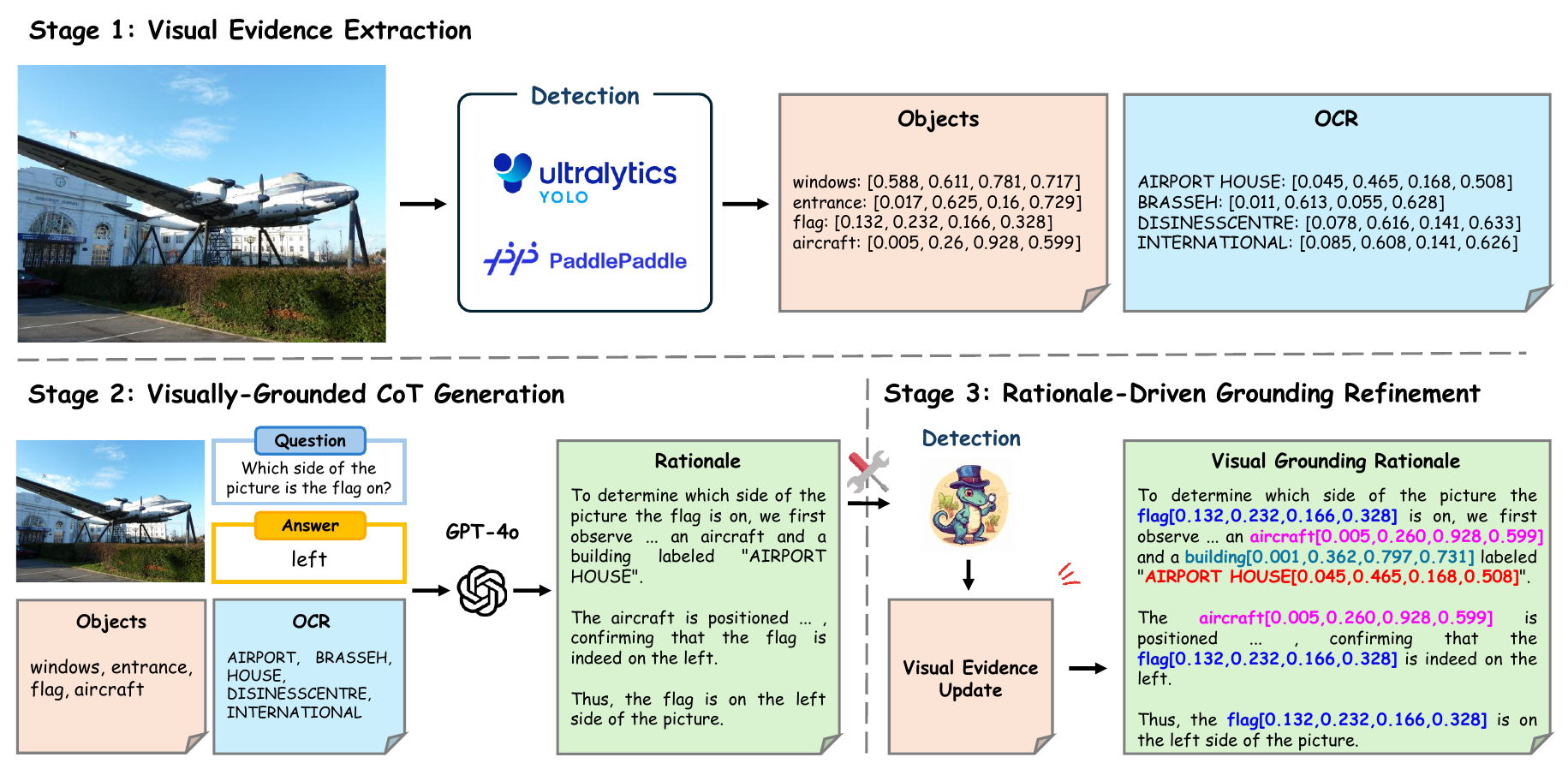}
    \caption{Overview of the Automated Pipeline for Generating Grounded CoT Data.}
\label{fig:figure2}
\end{figure*}

\subsection{Evaluation of Reasoning in LVLMs}

Existing benchmarks for evaluating the reasoning capabilities of LVLMs have traditionally been measured solely by the correctness of the model's final answer \cite{13_Antol_2015_ICCV, 14_Hudson_2019_CVPR}. This approach has a limitation: even if a model provides the correct answer, it cannot distinguish whether the result is due to rational reasoning based on correct visual evidence or merely due to statistical bias or hallucination \cite{39_Guan_2024_CVPR, 40_Agarwal_2020_CVPR, 41_Shah_2019_CVPR}. To overcome the limitations of this simple accuracy-based evaluation, new attempts have emerged to measure various capabilities and the reliability of models. MME \cite{22_fu2024mme} sought to comprehensively evaluate LVLMs across 14 fine-grained tasks by dividing the model's capabilities into perception and cognition. However, since it primarily focused on the accuracy of the final answer, it had the limitation of not evaluating the intermediate reasoning process that the model used to derive the answer or the visual evidence utilized during that process \cite{22_fu2024mme, 38_qiu-etal-2024-valor}.

In an effort to evaluate the reasoning process itself, CURE \cite{7_chen-etal-2024-measuring} was proposed to measure the consistency of CoT reasoning. It introduced a metric that provides high-level reasoning problems and step-by-step sub-questions to solve them, evaluating whether the model consistently answers not only the final correct answer but also the intermediate-step questions. While this was a significant advancement in evaluating the logical flow of the reasoning process, it failed to directly assess whether the model correctly referenced and utilized specific visual evidence within the image to solve the step-by-step sub-questions. Consequently, the current evaluation system still has a gap in comprehensively verifying the model's fine-grained reasoning ability in terms of `the logicality of the reasoning process' and `the utilization of visual evidence' \cite{21_jing-etal-2024-faithscore, 38_qiu-etal-2024-valor, 39_Guan_2024_CVPR}. Therefore, to fill this gap, this study proposes a new benchmark that comprehensively measures (1) Rationale Quality (visual evidence utilization, logicality, completeness), (2) Answer Accuracy, and (3) Reasoning-Answer Alignment to analyze the trustworthy reasoning capabilities of LVLMs.

\section{VG-CoT Dataset}
\label{sec:method}
\begin{table}[t]
\centering
\label{tab:vqa_rationale_prompt}
\begin{tabular}{p{0.95\linewidth}}
\Xhline{3\arrayrulewidth}
\toprule
\textbf{Prompt Template} \\ \hline
\midrule
\texttt{You are an AI assistant that generates step-by-step reasoning explanations (rationales) for visual question answering.} \\
\texttt{} \\
\texttt{\{object\_info\}\{ocr\_info\}} \\
\texttt{} \\
\texttt{Question: \{question\}} \\
\texttt{Answer: \{answer\}} \\
\texttt{} \\
\texttt{IMPORTANT GUIDELINES:} \\
\texttt{1. To solve the question, observe the image and write the reasoning process step-by-step.} \\
\texttt{2. Explicitly connect your reasoning process to the visual cues observed in the image.} \\
\texttt{3. Include reasoning about why certain things are NOT the answer or why other options don't apply.} \\
\texttt{4. Explain what makes the answer correct by comparing or contrasting with what is NOT true.} \\
\texttt{...} \\
\texttt{} \\
\texttt{Rationale:} \\
\Xhline{3\arrayrulewidth}
\bottomrule
\end{tabular}
\caption{Prompt Template for Generating step-by-step Rationales for VQA.}
\label{tab:table4}
\end{table}

\begin{figure*}[t]
    \centering
    \includegraphics[width=1.0\textwidth]{latex/figure/Figure_3.pdf}
    \caption{Detailed Statistics of the VG-CoT Dataset: Bounding Box Size Distribution (R: Relative Area Ratio) for Objects and OCR, and Key Metrics.}
\label{fig:figure3}
\end{figure*}

To overcome the limitations of existing datasets, this study introduces the Visual Grounding Chain-of-Thought (VG-CoT) dataset. The construction of VG-CoT is driven by three core objectives: (1) achieving scalability through a fully automated pipeline, (2) securing trustworthy reasoning by explicitly linking each logical step to precise visual evidence (i.e., spatial coordinates), and (3) encompassing diverse reasoning scenarios across multiple domains. As shown in Figure~\ref{fig:figure1}, each data sample systematically integrates an image, a question, an answer, and a step-by-step reasoning process densely grounded in multiple object and text locations.

\subsection{Data Generation Pipeline}
As illustrated in Figure~\ref{fig:figure2}, we develop a fully automated three-stage pipeline to construct the VG-CoT dataset without relying on expensive manual annotation. We source raw data from three distinct tasks to comprehensively cover the complex reasoning scenarios that LVLMs may encounter: Scene-text VQA (TextVQA) \cite{30_Singh_2019_CVPR} for text-intensive reasoning, General VQA (Visual7W) \cite{31_Zhu_2016_CVPR} for object recognition, and Relation VQA (GQA) \cite{14_Hudson_2019_CVPR} for complex inter-object relational reasoning.

\paragraph{Stage 1: Multi-Granularity Visual Evidence Extraction.}
The first stage extracts comprehensive visual evidence from the source images to serve as the foundational grounding data. To capture diverse visual elements, we apply specialized extractors for each modality:
\begin{itemize}
\item \textbf{Object-level Evidence:} The state-of-the-art YOLO model \cite{32_Redmon_2016_CVPR} detects the bounding boxes of key objects.
\item \textbf{Text-level Evidence:} PaddleOCR \cite{33_cui2025paddleocr} extracts the location and textual content within the image, robustly handling complex environments.
\end{itemize}
Additionally, for the GQA dataset, the rich scene graph information provided by the dataset itself is leveraged as supplementary initial evidence.

\paragraph{Stage 2: Visually-Grounded CoT Generation.}
The second stage bridges the gap between visual perception and logical reasoning. We provide GPT-4o, which possesses powerful multimodal capabilities, with the original image, the question, the ground-truth answer, and the visual evidence extracted in Stage~1. The model is explicitly instructed to synthesize this information and generate a step-by-step reasoning path. Crucially, the prompt enforces evidence-based reasoning by requiring the model to directly cite the spatial coordinates (visual evidence) obtained in Stage~1 for every logical deduction it makes toward the final answer.

\vspace{1em}
The specific instructions provided to GPT-4o for this generation task (detailed in Table~\ref{tab:table4}) ensure that the model utilizes the visual evidence fully, moving beyond simple text generation. Crucially, the prompt encourages the model to not only outline the path to the correct answer but also to analyze why certain elements are incorrect. This requires the model to demonstrate the plausibility of its answer through comparison and contrast against visual cues, resulting in a highly reliable and faithful VG-CoT.

\paragraph{Stage 3: Rationale-Driven Grounding Refinement.}
The final stage refines and extends the visual evidence based on the generated reasoning process. While the initial detection in Stage~1 extracts general objects and OCR text, the CoT generated in Stage~2 often references more subtle or context-dependent objects. To align these elements, we employ the following refinement process: First, key object nouns are extracted from the CoT sentences using natural language processing tools (e.g., nltk\footnote{\url{https://www.nltk.org/}}). These nouns then serve as queries for the open-set detector Grounding DINO~\cite{34_liu2025groundingdino}, enabling precise textual grounding of specific objects mentioned in the rationale. Finally, this refined visual evidence is explicitly linked to each reasoning step, ensuring clear alignment between the text and actual image coordinates.

Ultimately, this fully automated pipeline requires no human intervention, achieving high scalability and cost-efficiency while ensuring the consistent quality of the VG-CoT dataset by grounding each reasoning step in visual evidence across diverse scenarios.

\subsection{Data Analysis}

To comprehensively understand the characteristics of the VG-CoT dataset, this section provides a detailed analysis of its statistics, domain distribution, and visual evidence utilization patterns. By dissecting these aspects, we demonstrate that the dataset possesses rich, multi-dimensional features essential for training and evaluating the multifaceted reasoning capabilities of LVLMs. Figure~\ref{fig:figure3} summarizes the key statistics of the dataset, which comprises a total of 13,826 high-quality samples.

\vspace{0.5em}
\noindent\textbf{Dataset Scale and Task Distribution.}
The dataset spans three complementary tasks to cover diverse reasoning abilities, comprising General VQA (4,000 samples) for object understanding, Relation VQA (6,600 samples) for complex inter-object relationships, and Scene-text VQA (3,226 samples) for text comprehension. The total 13,826 samples are partitioned into training (10,226), validation (1,800), and test (1,800) splits at a 7:1.5:1.5 ratio.

\vspace{0.5em}
\noindent\textbf{Bounding Box Size Distribution.}
A distinctive feature of the VG-CoT dataset is the stark contrast in bounding box sizes across modalities. As illustrated in Figure~\ref{fig:figure3}, while object bounding boxes are predominantly medium or large, text bounding boxes are overwhelmingly tiny or small. This distribution creates a challenging and realistic learning environment, requiring models to precisely identify not only prominent objects but also fine-grained textual evidence.

\vspace{0.5em}
\noindent\textbf{Multi-Step Grounding Complexity.}
Unlike previous datasets limited to single-evidence referencing, the multi-step reasoning process in VG-CoT is explicitly linked to a vast amount of visual evidence. Specifically, the dataset incorporates 51,025 total object bounding boxes and 4,782 total OCR bounding boxes. This dense visual grounding is coupled with highly detailed logical steps, reflected by an average rationale length of 906.5. Such complexity provides an environment for training and evaluating the fine-grained, visually-grounded reasoning capabilities of LVLMs.

\subsection{Benchmark Construction}

To overcome the limitations of existing accuracy-centric evaluations, we establish a comprehensive benchmark designed to systematically assess the reasoning process of LVLMs. This benchmark is structured around three core dimensions.

\begin{itemize}
\item \textbf{Rationale Quality (RQ).}
We utilize a large language model (GPT-4o) as an automated evaluator to assess the quality of the reasoning process on a 1 to 5 scale. This dimension is comprehensively measured through four sub-metrics. \textbf{Visual Evidence} evaluates the degree to which the model leverages object and text location information during reasoning. \textbf{Coherence} assesses the logical flow between consecutive reasoning steps. \textbf{Completeness} measures whether all necessary logical deductions are present. Finally, the \textbf{Overall} score represents the combined average of these three aspects.

\item \textbf{Answer Accuracy (AA).} 
This metric determines whether the model's final prediction matches the ground truth. It represents the fundamental correctness of the generated output.

\item \textbf{Reasoning-Answer Alignment (RAA).} 
Going beyond independent metric evaluations, RAA measures the correlation between the intermediate reasoning quality and the final answer correctness. This allows us to verify how faithfully the reasoning process supports the final prediction. We define two detailed sub-metrics under this dimension.
\end{itemize}

\vspace{0.5em}
\noindent\textbf{Consistency Score.}
This sub-metric calculates the proportion of samples where the rationale quality logically aligns with the answer correctness. A match is defined as either providing excellent reasoning ($RQ \ge 4$) for a correct answer ($AA = 1$) or poor reasoning ($RQ \le 3$) for an incorrect answer ($AA = 0$).
\begin{equation}
\text{Consistency Score} = \frac{|\mathcal{S}_{\text{consistent}}|}{N}
\end{equation}
Where:
\begin{itemize}
\item $\mathcal{S}_{\text{consistent}} = \{ i : (RQ_i \ge 4 \land AA_i = 1) \lor (RQ_i \le 3 \land AA_i = 0) \}$
\item $N$ is the total number of evaluated samples.
\item $i$ is the index of an individual data sample.
\item $AA_i \in \{0, 1\}$ denotes the binary correctness of the final answer.
\end{itemize}

\vspace{0.5em}
\noindent\textbf{Faithful Score.}
This sub-metric evaluates the reliability of the model's correct predictions by measuring the proportion of correct answers strictly among the samples that exhibit excellent reasoning ($RQ \ge 4$).
\begin{equation}
\text{Faithful Score} = \frac{|{ i : RQ_i \ge 4 \land AA_i = 1 }|}{|{ i : RQ_i \ge 4 }|}
\end{equation}

\section{Experiment}
\label{sec:experiment}
% \usepackage{multirow}
\begin{table*}[t]
\renewcommand{\arraystretch}{1.5} % Adjust this number to change row height
\centering
\resizebox{1.0\textwidth}{!}{%
\begin{tabular}{l|cccc|c|cc}
\Xhline{4\arrayrulewidth}
\multicolumn{1}{c|}{}                                 & \multicolumn{4}{c|}{\textbf{Rationale Quality}}                                          &                                                                                      & \multicolumn{2}{c}{\textbf{\begin{tabular}[c]{@{}c@{}}\makecell{\textbf{Reasoning-Answer}\\\textbf{Alignment}}\end{tabular}}} \\ \cline{2-5} \cline{7-8} 
\multicolumn{1}{c|}{\multirow{-2}{*}{\textbf{Model}}} & \textbf{Visual Evidence} & \textbf{Coherence} & \textbf{Completeness} & \textbf{Overall} & \multirow{-2}{*}{\textbf{\begin{tabular}[c]{@{}c@{}}\makecell{\textbf{Answer}\\\textbf{Accuracy}}\end{tabular}}} & \textbf{Consistency}                              & \textbf{Faithful}                             \\ \hline\hline
LLaVA-1.5-7B                                          & 67.2                     & 77.8               & 71.6                  & 72.2             & 48.7                                                                                 & 60.9                                              & 58.9                                          \\
\rowcolor[HTML]{EDEBEB} 
\textbf{+ VG-CoT}                                     & \textbf{79.4}            & \textbf{83.9}      & \textbf{87.0}         & \textbf{83.4}    & \textbf{62.5}                                                                        & \textbf{64.1}                                     & \textbf{64.3}                                 \\ \hline
LLaVA-1.5-13B                                         & 71.3                     & 83.1               & 74.3                  & 76.2             & 58.2                                                                                 & \textbf{66.9}                                     & \textbf{69.7}                                 \\
\rowcolor[HTML]{EDEBEB} 
\textbf{+ VG-CoT}                                     & \textbf{79.8}            & \textbf{94.3}      & \textbf{87.7}         & \textbf{87.3}    & \textbf{63.2}                                                                        & 65.6                                              & 66.2                                          \\ \hline
Qwen2-VL-7B-Instruct                                           & 79.1                     & 90.8               & 84.1                  & 84.7             & 64.9                                                                                 & 66.7                                              & 69.1                                          \\
\rowcolor[HTML]{EDEBEB} 
\textbf{+ VG-CoT}                                     & \textbf{81.2}            & \textbf{94.8}      & \textbf{88.9}         & \textbf{88.3}    & \textbf{72.3}                                                                        & \textbf{72.9}                                     & \textbf{75.0}                                 \\ \hline
Qwen2.5-VL-7B-Instruct                                         & 80.9                     & 94.9               & 88.3                  & 88.0             & 68.5                                                                                 & 68.5                                              & 70.0                                          \\
\rowcolor[HTML]{EDEBEB} 
\textbf{+ VG-CoT}                                     & \textbf{82.9}            & \textbf{95.0}      & \textbf{90.6}         & \textbf{89.5}    & \textbf{73.6}                                                                        & \textbf{72.6}                                     & \textbf{75.7}  \\ \Xhline{4\arrayrulewidth}                              
\end{tabular}%
}
\caption{Performance Comparison on the Proposed VG-CoT Benchmark.}
\label{tab:tab1}
\end{table*}

\begin{table*}[t]
\renewcommand{\arraystretch}{1.5} % Adjust this number to change row height
\centering
\resizebox{0.9\textwidth}{!}{%
\begin{tabular}{l|cc|cc|cc}
\Xhline{3\arrayrulewidth}
\multicolumn{1}{c|}{}                                 & \multicolumn{2}{c|}{\textbf{General VQA}}                                   & \multicolumn{2}{c|}{\textbf{Relation VQA}}                                  & \multicolumn{2}{c}{\textbf{Scene-text}}                                     \\ \cline{2-7} 
\multicolumn{1}{c|}{\multirow{-2}{*}{\textbf{Model}}} & \textbf{Rationale}                   & \textbf{Answer}                             & \textbf{Rationale}                            & \textbf{Answer}                              & \textbf{Rationale}                            & \textbf{Answer}                             \\ \hline\hline
LLaVA-1.5-7B                                          & 71.8                                 & 51.8                                 & 69.8                                 & 53.4                                 & 63.8                                 & 38.0                                 \\
\rowcolor[HTML]{EDEBEB} 
\textbf{+ VG-CoT}                                      & \textbf{86.4}                        & \textbf{69.2}                        & \textbf{84.1}                        & \textbf{64.3}                        & \textbf{86.4}                        & \textbf{43.0}                        \\ \hline
LLaVA-1.5-13B                                         & 76.4                                 & 74.0                                 & 73.6                                 & 59.3                                 & 67.7                                 & 40.8                                 \\
\rowcolor[HTML]{EDEBEB} 
\textbf{+ VG-CoT}                                      & \textbf{87.9}                        & \textbf{79.8}                        & \textbf{85.1}                        & \textbf{64.4}                        & \textbf{86.2}                        & \textbf{44.8}                        \\ \hline
Qwen2-VL-7B-Instruct                                           & 82.0                                 & 70.4                                 & 81.9                                 & 60.5                                 & 84.9                                 & 66.6                                 \\
\rowcolor[HTML]{EDEBEB} 
{\color[HTML]{000000} \textbf{+ VG-CoT}}               & {\color[HTML]{000000} \textbf{89.0}} & {\color[HTML]{000000} \textbf{81.8}} & {\color[HTML]{000000} \textbf{86.0}} & {\color[HTML]{000000} \textbf{65.6}} & {\color[HTML]{000000} \textbf{88.3}} & {\color[HTML]{000000} \textbf{73.6}} \\ \hline
Qwen2.5-VL-7B-Instruct                                & 88.5                                 & 79.4                                 & 84.7                                 & 57.8                                 & 88.2                                 & 74.2                                 \\
\rowcolor[HTML]{EDEBEB} 
\textbf{+ VG-CoT}                                      & \textbf{89.9}                        & \textbf{84.4}                        & \textbf{88.2}                        & \textbf{66.4}                        & \textbf{90.6}                        & \textbf{74.8} \\
\Xhline{3\arrayrulewidth}
\end{tabular}%
}
\caption{Task-Specific LVLMs Performance Comparison: Effect of VG-CoT Fine-Tuning.}
\label{tab:tab2}
\end{table*}

In this section, we conduct experiments to validate the effectiveness of the proposed VG-CoT dataset and demonstrate the utility of our new benchmark.

\subsection{Experimental Setup}

\noindent\textbf{Evaluated LVLMs.}
To validate the effectiveness of our proposed dataset, we perform fine-tuning and evaluation on four representative LVLMs: LLaVA-1.5 (7B and 13B) \cite{24_Liu_2024_CVPR}, Qwen2-VL-7B-Instruct \cite{26_wang2024qwen2}, and Qwen2.5-VL-7B-Instruct \cite{29_bai2025qwen2}.

\vspace{0.5em}\noindent\textbf{Benchmarks.}
To comprehensively evaluate the validity of the models' reasoning capabilities, we utilize our proposed benchmark rather than relying solely on traditional accuracy metrics. The models are evaluated across three core dimensions. First, Rationale Quality (RQ) assesses visual evidence utilization, logical coherence, and reasoning completeness. Second, Answer Accuracy (AA) represents the overall correctness of the model's output. Finally, Reasoning-Answer Alignment (RAA) measures the consistency and faithfulness between the generated rationale and the final prediction.

\vspace{0.5em}\noindent\textbf{Implementation Details.}
All comparative models are fine-tuned using the VG-CoT dataset. During the training phase, the models take an image and a question as input, and are optimized to generate a step-by-step reasoning process explicitly grounded in visual evidence, followed by the final answer. All experiments are conducted in an environment equipped with two NVIDIA H100 (80GB) GPUs. To ensure computational efficiency, we employ Low-Rank Adaptation (LoRA) \cite{35_hu2022lora} for Parameter-Efficient Fine-Tuning (PEFT). The training process is executed for a total of 3 epochs, utilizing a learning rate of 2e-4 and a batch size of 32.

\subsection{Overall Performance on VG-CoT Benchmark}

Table~\ref{tab:tab1} presents the performance analysis of four representative LVLMs evaluated on the proposed VG-CoT benchmark. Overall, fine-tuning with the VG-CoT dataset yielded improvements across most dimensions. All evaluated models exhibited consistent gains in both Rationale Quality (RQ) and Answer Accuracy (AA). For instance, the overall RQ of LLaVA-1.5-7B increased from 72.2 to 83.4, while its AA rose from 48.7 to 62.5. This confirms that explicitly linking the reasoning process to visual evidence effectively enhances trustworthy and logical reasoning capabilities.

Despite these overall gains, specific alignment anomalies and grounding bottlenecks were observed. LLaVA-1.5-13B showed a slight decrease in Consistency from 66.9 to 65.6, implying an occasional disconnect between generated rationales and final answers. Furthermore, the Visual Evidence score remained consistently lower than other RQ components across all models. This indicates that precise spatial grounding is intrinsically more challenging than logical text generation, a trend further supported by the results in Table~\ref{tab:tab3}.

\subsection{Task-Specific Fine-Tuning Effects and Validity}

Table~\ref{tab:tab2} shows the effect of VG-CoT fine-tuning across three distinct reasoning tasks. The results demonstrate that the proposed dataset consistently enhances both Rationale Quality and Answer Accuracy across all evaluated domains, enabling all models to exhibit significant improvements in General and Relation VQA tasks. Most notably, LLaVA-1.5-7B demonstrated remarkable gains of over 10 points in both reasoning and correctness, with its Answer Accuracy jumping from 51.8 to 69.2 in General VQA and from 53.4 to 64.3 in Relation VQA. This confirms that VG-CoT effectively trains models to reason logically about spatial and relational contexts.

The benefits of VG-CoT extend to text-heavy reasoning and already highly capable models. In the challenging Scene-text task, the Rationale score of LLaVA-1.5-7B surged from 63.8 to 86.4. Furthermore, advanced instruction-tuned models like Qwen2.5-VL-7B-Instruct also saw substantial boosts, particularly in General VQA Answer Accuracy, which increased from 79.4 to 84.4. This task-agnostic and consistent improvement comprehensively validates the broad applicability of the VG-CoT dataset.

\begin{table}[t]
\renewcommand{\arraystretch}{1.5} % Adjust this number to change row height
\centering
\resizebox{1\columnwidth}{!}{%
\begin{tabular}{c|cc|cc}
\Xhline{4\arrayrulewidth}

\multirow{2}{*}{}     & \multicolumn{2}{c|}{\textbf{LLaVA-1.5-7B}} & \multicolumn{2}{c}{\textbf{LLaVA-1.5-13B}} \\ \cline{2-5} 
                      & \textbf{mAP@0.5}    & \textbf{mAP@0.75}    & \textbf{mAP@0.5}    & \textbf{mAP@0.75}    \\ \hline\hline
\textbf{General VQA}  & 46.65               & 29.07                & 52.2                & 36.4                 \\
\textbf{Relation VQA} & 50.16               & 29.84                & 54.9                & 35.7                 \\
\textbf{Scene-text}   & 30.93               & 21.64                & 32.4                & 26.3                 \\
\textbf{Average}      & 44.5                & 27.64                & 48.7                & 33.6  \\
\Xhline{4\arrayrulewidth}

\end{tabular}%
}
\caption{Accuracy of Predicted Visual Evidence for VG-CoT Fine-tuned Models.}
\label{tab:tab3}
\end{table}

\subsection{Analysis of Visual Evidence Prediction Accuracy}

Table~\ref{tab:tab3} details the visual evidence prediction accuracy of the fine-tuned LLaVA models using mean Average Precision (mAP) for bounding box localization against pseudo-label ground truths. As a standard metric for object detection, mAP evaluates the accuracy of predicted bounding boxes by measuring their overlap with ground truth regions at specific Intersection over Union (IoU) thresholds. The results indicate that increased model capacity enhances grounding ability, with LLaVA-1.5-13B outperforming the 7B variant in average mAP@0.5 by achieving 48.7 compared to 44.5. Across tasks, models achieved their highest scores in Relation VQA, reaching 54.9 mAP@0.5 for the 13B model, but struggled significantly in the Scene-text task with a score of 32.4. This substantial gap quantitatively demonstrates that while models can reliably ground larger relational objects, precisely locating small and dense text regions remains intrinsically challenging.

Furthermore, the data reveals a steep performance drop when applying a stricter IoU threshold. For instance, the average precision of LLaVA-1.5-13B drops from 48.7 at mAP@0.5 to 33.6 at mAP@0.75. This highlights that although the VG-CoT fine-tuned models successfully learn to approximate the general location of relevant visual evidence, achieving highly precise spatial grounding is still a significant bottleneck that requires further exploration.

\section{Conclusion}
\label{sec:conclusion}
This study aimed to address the limitations of existing datasets and benchmarks to enhance the trustworthy rationale-based reasoning capabilities of LVLMs. To this end, we proposed the VG-CoT dataset and an automated three-stage pipeline for its construction, enabling the scalable and reliable generation of visual evidence-based CoT data. Additionally, we established a new benchmark that comprehensively measures Rationale Quality, Answer Accuracy, and Reasoning-Answer Alignment to deeply evaluate the model's reasoning process. Experimental results confirmed that VG-CoT consistently improved reasoning quality and answer accuracy, and also showed improvements in consistency for most models. Furthermore, an analysis of the model's prediction accuracy for visual evidence suggested the potential for improving grounding capability along with future challenges. This study provides a core dataset and evaluation methodology for trustworthy LVLMs reasoning and is expected to contribute to the advancement of related research in the future.

\section{Limitations}
Although this study contributes to enhancing visual evidence-based reasoning in LVLMs, it has certain limitations that suggest directions for future work.

First, as the proposed VG-CoT framework is a fully automated three-stage pipeline designed for scalability, its performance is inherently tied to the integration of its underlying foundational models, such as YOLO, PaddleOCR, Grounding DINO and GPT-4o. While this cohesive approach enables cost-efficient dataset construction, the final quality remains subject to the precision of these models. Furthermore, while we focused on validating the overall effectiveness and scalability of the integrated system, a more granular analysis to disentangle the isolated impact of each stage would provide deeper insights into the grounding mechanism.

Second, while the VG-CoT dataset covers a wide range of general and relational reasoning tasks, its applicability to highly specialized professional domains, such as medical diagnostics or engineering diagram comprehension, has yet to be fully explored. Expanding the pipeline to these specific fields remains a promising direction for future research.

\section{Ethics Statement}
This study was conducted using publicly available datasets (GQA, Visual7W, TextVQA) that contain no personally identifiable or sensitive information. The proposed automated pipeline does not involve human annotation or data collection from individuals, thereby minimizing ethical risks. All experiments and analyses were performed in compliance with the ethical standards of dataset usage and research transparency guidelines.

\section{Acknowledgements}
This work was supported by the Institute of Information \& Communications Technology Planning \& Evaluation (IITP) grant funded by the Korea government (MSIT) [RS-2021-II211341, Artificial Intelligence Graduate School Program (Chung-Ang University)] and by the National Research Foundation of Korea (NRF) grant funded by the Korea government (MSIT) (RS-2025-00556246).

\section{Bibliographical References}\label{sec:reference}

\bibliographystyle{lrec2026-natbib}
\bibliography{latex/lrec2026}

% \section{Language Resource References}
\label{lr:ref}
\bibliographystylelanguageresource{lrec2026-natbib}
\bibliographylanguageresource{languageresource}

\end{document}